\documentclass[11pt]{article}

\usepackage[preprint]{acl}

\usepackage{times}
\usepackage{latexsym}

\usepackage[T1]{fontenc}
\usepackage[utf8]{inputenc}

\usepackage{microtype}
\usepackage{inconsolata}

\usepackage{graphicx}
\usepackage{multirow}
\usepackage{subcaption}
\usepackage{amsmath}
\usepackage{colortbl}

\usepackage{xcolor}
\definecolor{BLUE_SHADE}{HTML}{3285a8}
\definecolor{POSSIBLE_SHADE}{HTML}{0a089c}

\usepackage{hyperref} 

\definecolor{tablegray}{gray}{0.80}
\newcommand{\groupline}{%
  \arrayrulecolor{tablegray}\hline
  \arrayrulecolor{black}%
}

\title{\textbf{\textcolor{POSSIBLE_SHADE}{SuPER-Mamba:} \textcolor{BLUE_SHADE}{Surprisal Prioritization for Extended-Context Routing in Mamba}}}

\author{
  Zhetong Zhang \and Gayan K Kulatilleke\thanks{~Corresponding author.} \\
  School of Information Technology and Electrical Engineering \\
  The University of Queensland \\
  \texttt{zhetong.zhang@student.uq.edu.au, g.kulatilleke@uq.edu.au}
}

\begin{document}

\maketitle


\begin{abstract}
State space models provide efficient linear-complexity processing for long sequences, but their performance frequently degrades as context length grows. We propose SuPER-Mamba, a training-free framework that uses Surprisal as a model-intrinsic signal for long-range state routing in pretrained Mamba models. SuPER-Mamba computes position-wise Surprisal from next-token predictions, aggregates neighboring responses through Local Pooling, and uses the resulting scores to prioritize Global-Channel updates without query-dependent scoring or additional training. We evaluate SuPER-Mamba on three state-space models across the LongBench-E benchmark. The results show that SuPER-Mamba improves the average score from 8.06 to 15.69, corresponding to a 94.7\% relative gain over the Base Models and a 13.1\% improvement over the state-of-the-art prior baseline. Ablation and length-wise analyses further support the effectiveness of the proposed routing framework.
\end{abstract}

\noindent {Code, training scripts and datasets : \url{https://github.com/gayanku/SuPER-Mamba}

\begin{figure}[t]
    \centering
    \includegraphics[width=\columnwidth]{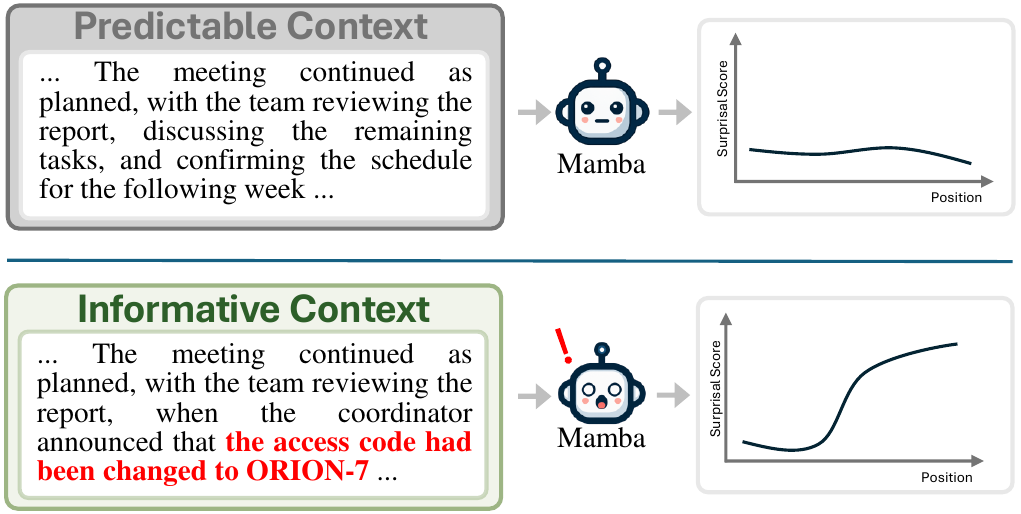}
    \caption{Motivation of SuPER-Mamba. A pretrained Mamba produces position-wise Surprisal from next-token prediction probabilities while processing the input context. Relatively predictable local content typically produces lower and more stable Surprisal responses, whereas less predictable local changes may produce larger responses. SuPER-Mamba treats these responses as model-intrinsic predictive feedback and further investigates the relationship between such local predictive feedback and signals for long-range state allocation.}
    \label{fig:motivation}
\end{figure}



\section{Introduction}

Long-context modeling requires a model to process large amounts of input while effectively preserving and utilizing information that remains relevant over long distances. Mamba and its subsequent variants provide an efficient alternative to Transformer-based architectures for long-sequence modeling through recurrent state-space computation with linear sequence complexity~\citep{gu-dao-2023-mamba,dao-gu-2024-mamba2}. However, historical information must be continuously compressed into a finite recurrent state, making effective long-range information preservation increasingly difficult as the context grows. Existing studies have shown that different hidden channels in Mamba exhibit different long-range retention characteristics, and that selectively controlling long-range state updates can alleviate information decay and improve long-context modeling performance~\citep{ye-etal-2025-longmamba,ye-etal-2025-lamb}. These findings regarding Mamba's long-context capabilities motivate the investigation of intrinsic model signals that can guide the allocation of limited long-range state updates without relying on downstream queries.

Pretrained autoregressive language models continuously produce predictive distributions over the next token while processing a context, reflecting the model's predictive expectations at different positions. As illustrated in Figure~\ref{fig:motivation}, relatively predictable local content tends to produce more stable predictive responses, whereas less predictable changes may induce more pronounced Surprisal responses. The mismatch between the model's predictive expectation and the observed next token motivates our hypothesis that these position-dependent predictive responses can provide a model-intrinsic signal for long-range state allocation.

Based on this hypothesis, we propose SuPER-Mamba, which uses Surprisal derived from the complete input context to quantify position-wise predictive responses and combines them with local information aggregation to construct context scores for guiding long-range state updates. Figure~\ref{fig:framework} provides an overview of the framework. SuPER-Mamba first computes position-wise Surprisal and aggregates neighboring responses through Local Pooling, then uses the resulting scores to route selected content positions to long-range state updates while preserving the complete input in the original recurrent computation. The method requires no additional training or model fine-tuning. We systematically evaluate SuPER-Mamba on 13 LongBench-E tasks across three representative state-space-based architectures. The results show that, without using the downstream query for context selection, SuPER-Mamba substantially improves the long-context performance of the corresponding Base Models while remaining competitive with state-of-the-art baselines.

Our contributions are as follows.

\begin{itemize}

    \item We propose SuPER-Mamba, a query-independent routing framework for long-context modeling in pretrained Mamba models. The framework directly uses predictive responses produced by the pretrained model to construct context-scoring signals to guide long-range state updates.

    \item We first introduce Surprisal-guided routing for long-context modeling in pretrained Mamba models. SuPER-Mamba combines Surprisal with local information aggregation to transform the model's predictive feedback into a routing signal for long-range state allocation without additional selector training.

    \item We evaluate SuPER-Mamba on 13 LongBench-E tasks across three representative state-space-based architectures. Without using the downstream query for context selection, SuPER-Mamba improves the average score from 8.06 to 15.69, corresponding to a 94.7\% relative improvement over the Base Models, while achieving 13.1\% higher average performance than the state-of-the-art baseline evaluated across the same architectures.

\end{itemize}

\begin{figure*}[t]
    \centering
    \includegraphics[width=0.98\textwidth]{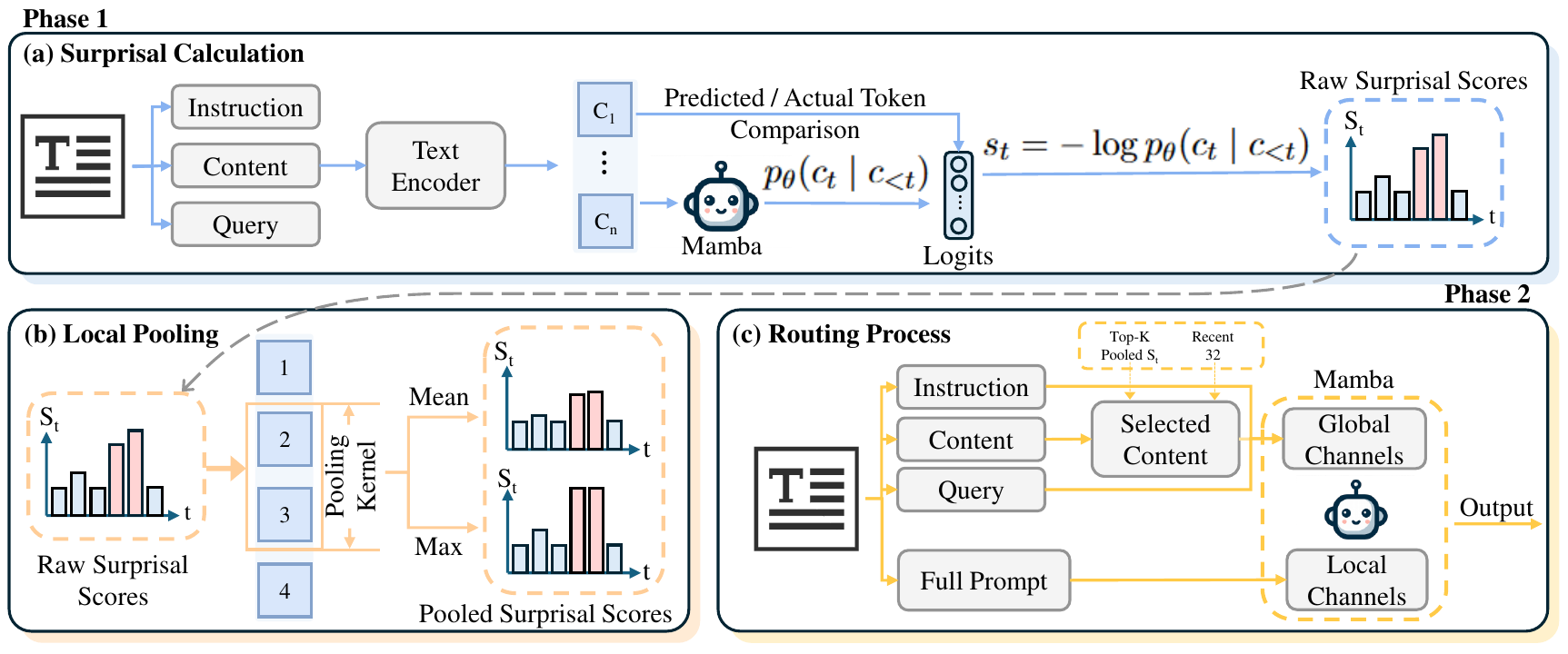}
    \caption{Overview of the SuPER-Mamba framework. Phase 1 performs offline Context Scoring. (a) Surprisal Calculation computes position-wise Raw Surprisal Scores from the next-token prediction probabilities of the pretrained Mamba. (b) Local Pooling aggregates neighboring predictive responses using category-specific Mean or Max Pooling to obtain Pooled Surprisal Scores. Phase 2 uses these scores for long-range state routing. The historical Top-$K$ Content positions selected by the Pooled Surprisal Scores are combined with the most recent $R=32$ Content positions to form the final routing set. For long-range state updates, Instruction and Query positions do not participate in Content Selection and retain their original state updates, while among Content positions, only the selected positions update the predefined Global Channels. All Content positions continue to participate in the original recurrent computation, so the complete input is neither removed nor truncated.}
    \label{fig:framework}
\end{figure*}

\begin{table*}[t]
\centering
\small
\setlength{\tabcolsep}{4.5pt}
\renewcommand{\arraystretch}{1.18}

\begin{tabular}{llcccc}
\hline
\textbf{Category}
& \textbf{Dataset}
& \textbf{Base Mamba2}
& \textbf{LongMamba}
& \textbf{LAMB}
& \textbf{SuPER-Mamba (Ours)} \\
\hline

\multirow{2}{*}{Single-doc QA}
& Qasper
& 1.76
& 4.75
& \textbf{5.63}
& \underline{5.00} \\

& MultiFieldQA-en
& 5.03
& 11.12
& \textbf{14.76}
& \underline{13.81} \\
\groupline

\multirow{2}{*}{Multi-doc QA}
& HotpotQA
& 2.89
& 5.45
& \textbf{7.71}
& \underline{6.43} \\

& 2WikiMQA
& 2.31
& 5.19
& 8.01
& \textbf{10.09} \\
\groupline

\multirow{2}{*}{Summarization}
& GovReport
& 7.66
& 14.67
& \textbf{16.38}
& 13.70 \\

& MultiNews
& 11.31
& 14.72
& \textbf{17.49}
& 14.45 \\
\groupline

\multirow{3}{*}{Few-shot}
& TREC
& 14.67
& 21.33
& \textbf{46.04}
& \underline{40.33} \\

& TriviaQA
& 10.08
& 47.58
& \textbf{51.04}
& \underline{47.69} \\

& SAMSum
& 4.66
& 13.79
& 26.02
& \textbf{28.30} \\
\groupline

\multirow{2}{*}{Synthetic}
& Passage Count
& 1.29
& 1.72
& 3.79
& \textbf{4.00} \\

& Passage Retrieval-en
& 0.81
& \textbf{5.60}
& 4.30
& \underline{5.19} \\
\groupline

\multirow{2}{*}{Code}
& LCC
& 22.85
& 40.39
& 41.65
& \textbf{49.03} \\

& RepoBench-P
& 19.89
& 33.52
& 34.02
& \textbf{42.20} \\
\hline

\multicolumn{2}{l}{\textbf{Average}}
& 8.09
& 16.91
& 21.30
& \textbf{21.56} \\
\hline

\end{tabular}
\caption{Main LongBench-E results on Mamba2-1.3B. Results are grouped by task category. Different datasets are evaluated using their official LongBench-E task-specific metrics, including QA F1 for Qasper, MultiFieldQA-en, HotpotQA, 2WikiMQA, and TriviaQA, ROUGE-L F1 for GovReport, MultiNews, and SAMSum, classification score for TREC, count score for Passage Count, retrieval score for Passage Retrieval-en, and code similarity for LCC and RepoBench-P. Each dataset score is the arithmetic mean over the 0--4k, 4--8k, and 8k+ context-length ranges. The Average row reports the arithmetic mean over all 13 dataset-level scores. Best results are shown in \textbf{bold}, and second-best results are \underline{underlined}.}
\label{tab:main-results}
\end{table*}

\section{Related Work}
\label{sec:related-work}

\subsection{Long-Context Memory and Routing}
\label{sec:related-mamba}

Mamba achieves linear-complexity sequence modeling through input-dependent state-space updates, but its fixed-size recurrent state requires growing historical information to be compressed into a bounded representation~\citep{gu-dao-2023-mamba,dao-gu-2024-mamba2}. Recent studies have examined how information is propagated and retained under this recurrent computation. The Hidden Attention of Mamba Models analyzes interactions between context positions from an attention-like perspective, while LaTIM characterizes fine-grained latent token-to-token interactions in Mamba-1 and Mamba-2~\citep{hidden-attention-mamba-2025,latim-2025}. Other work reports selective information loss in finite recurrent representations and shows that Mamba handles locally concentrated key information more effectively than information distributed across distant positions~\citep{selective-memory-mamba-2025,local-pattern-shortcuts-mamba-2025}. These findings indicate that long-context performance depends not only on input length, but also on how historical information is preserved within limited recurrent states.

Recent work has improved Mamba's long-context capability through state-dynamics modulation, selective compression, training-based memory enhancement, hierarchical historical access, and additional memory structures~\citep{mamba-modulation-2025,remamba-2025,recall-reasoning-mamba-2025,hierarchical-sparse-attention-2025,mamba-hierarchical-memory-2026,mamba3-2026,collapse-to-control-2026}. LongMamba and LAMB instead focus directly on selective long-range state updates. LongMamba identifies hidden channels with different receptive-field characteristics and applies selective token filtering to the channels responsible for longer-range information propagation~\citep{ye-etal-2025-longmamba}. LAMB follows this framework and derives attention-guided importance scores from Mamba's attention-like representation to determine which positions should update long-range states~\citep{ye-etal-2025-lamb}. These methods establish selective allocation of long-range updates as an effective training-free direction for pretrained Mamba models.

\subsection{Content-Driven Information Selection for Long Contexts}
\label{sec:related-selection}

Beyond recurrent state-space models, long-context research increasingly treats historical information as a non-uniform resource-allocation problem. H$_2$O, Scissorhands, SnapKV, PyramidKV, FastGen, and AttentionPredictor retain or predictively select historical KV states using attention statistics, persistence of importance, prompt observations, head-specific behavior, or temporal attention patterns~\citep{zhang-etal-2023-h2o,scissorhands-2023,li-etal-2024-snapkv,cai-etal-2024-pyramidkv,fastgen-2024,attentionpredictor-2025}. More recent work also considers local structure rather than isolated token scores. ChunkKV performs selection over contiguous chunks to better preserve local semantic structure, while DAST allocates compression capacity according to segment-level information density and combines local predictive information with global signals~\citep{chunkkv-2025,dast-2025}.

A related line of work reduces long inputs through context selection or prompt compression. Selective Context uses language-model self-information to remove predictable lexical units, while LLMLingua and LongLLMLingua use model probabilities, token-level information, and downstream questions for fine-grained compression~\citep{li-etal-2023-compressing,jiang-etal-2023-llmlingua,jiang-etal-2024-longllmlingua}. Other methods use task instructions, query relevance, user intention, answerability, or reasoning states to determine which passages, KV blocks, or evidence should be retained~\citep{perception-compressor-2025,adaptive-k-context-selection-2025,intentkv-2026,itercomp-2026}. Analyses of prompt compression further show that reducing input can discard contextual details that are difficult to recover~\citep{prompt-compression-information-preservation-2025}. These approaches therefore differ both in their scoring signals and in whether they operate on explicit KV states, modify the input, or condition selection on downstream information needs.

\subsection{Surprisal and Predictive Feedback for Context Selection}
\label{sec:related-surprisal}

Surprisal is an information-theoretic quantity defined from conditional prediction probabilities and has been used to characterize contextual predictability in incremental language processing~\citep{hale-2001-probabilistic,levy2008expectation}. For pretrained autoregressive models, it can be obtained directly from the next-token predictive distribution produced during context processing, providing a position-dependent response conditioned on the input and model state. Importantly, Surprisal measures predictive difficulty rather than semantic importance for a specific downstream task.

Predictive information has already been used for context selection. Selective Context uses self-information to distinguish more predictable context units, the LLMLingua family incorporates model probabilities or perplexity into compression decisions, and DAST combines local predictive information with global signals~\citep{li-etal-2023-compressing,jiang-etal-2023-llmlingua,jiang-etal-2024-longllmlingua,dast-2025}. However, predictive uncertainty and task-specific relevance are not equivalent. QUITO-X argues from an information-bottleneck perspective that high self-information or perplexity does not necessarily imply greater relevance to a downstream query~\citep{quitox-2025}. Related work on information preservation in prompt compression similarly cautions against treating prediction-derived scores as direct semantic-importance labels~\citep{prompt-compression-information-preservation-2025}.

Position-wise predictive responses can also vary because of lexical choice, special symbols, or local expression changes. This motivates considering local structure rather than isolated Surprisal peaks. Related work such as ChunkKV and DAST similarly suggests that contiguous or locally structured information can provide a more stable basis for selection than independent token-level scores~\citep{chunkkv-2025,dast-2025}.

Taken together, prior work shows that Mamba long-context modeling depends on information preservation in finite recurrent states, that selective long-range updates can improve performance, and that pretrained models already provide predictive signals for differentiating context positions. SuPER-Mamba connects these directions by preserving the complete input, deriving position-wise predictive feedback from Surprisal, aggregating neighboring responses through Local Pooling, and using the resulting context scores to allocate long-range state updates without additional selector training or downstream-query-dependent context scoring.

\section{Method}
\label{sec:method}

SuPER-Mamba constructs routing scores from the predictive feedback produced by a pretrained model during context processing. Given a long context, Surprisal first converts next-token predictions into position-wise predictive responses. Local information aggregation then combines neighboring responses into context scores, which are used to determine which positions update the predefined Global Channels.

The context-scoring procedure consists of Surprisal estimation and local information aggregation, followed by long-range state routing. Figure~\ref{fig:framework} illustrates the complete procedure.

\subsection{Surprisal Based Contextual Prediction}
\label{sec:surprisal}

Given a context
\begin{equation}
    C = (c_1, c_2, \ldots, c_N),
\end{equation}
where $c_t$ denotes the token at context position $t$, an autoregressive language model assigns a conditional probability to $c_t$ based on the preceding sequence $c_{<t}$,
\begin{equation}
    p_{\theta}(c_t \mid c_{<t}),
\end{equation}
where $\theta$ denotes the pretrained model parameters. The Surprisal at position $t$ is defined as
\begin{equation}
    s_t =
    -\log p_{\theta}(c_t \mid c_{<t}).
\end{equation}

Surprisal characterizes the unpredictability of the current token given its preceding context. A lower conditional probability corresponds to a higher Surprisal value and therefore indicates a stronger predictive response at the corresponding position. Conversely, when the current token is more consistent with the prediction formed from the preceding context, its conditional probability increases and the corresponding Surprisal decreases.

This formulation converts the predictive feedback continuously produced during autoregressive processing into an explicit position wise numerical signal. For a context of length $N$, the Surprisal values across all positions form the response sequence
\begin{equation}
    \mathbf{s}
    =
    (s_1, s_2, \ldots, s_N).
\end{equation}

The sequence $\mathbf{s}$ describes how the predictive response of the model varies across context positions. Different positions therefore receive differentiated response values according to the next token distributions produced by the pretrained model. Since these values are derived directly from the predictive distributions of the model, they provide input dependent position wise feedback without requiring a separate scoring model.

Surprisal nevertheless represents the predictive outcome at an individual position, and its values may vary substantially across neighboring tokens. Individual tokens that are difficult to predict may also produce isolated high responses. The Surprisal sequence therefore serves as the position wise predictive representation from which subsequent local information aggregation constructs the context scores used for routing.

\subsection{Local Information Aggregation}
\label{sec:aggregation}

Token-level Surprisal reflects the predictive response at an individual position and may vary substantially across neighboring tokens. Isolated high responses can arise from local lexical changes, special symbols, or other position-specific factors. Directly using these responses for routing may therefore produce unstable position selection.

Given the Surprisal sequence $\mathbf{s}$, we apply a local pooling operator
\begin{equation}
    \mathbf{u}
    =
    P_k(\mathbf{s}),
\end{equation}
where $P_k$ denotes Mean or Max pooling with window size $k$. The resulting score at each position incorporates predictive responses from its local neighborhood, reducing the influence of isolated peaks while preserving local response patterns. The aggregated sequence $\mathbf{u}$ is subsequently used for long-range state routing.

\subsection{Long-Range State Routing}
\label{sec:routing}

After local aggregation, the pooled score sequence is used to determine which context positions are allowed to update the predefined Global Channels. Let $R$ denote the recent-position window and let $\mathbf{u}$ denote the locally aggregated score sequence. SuPER-Mamba excludes the most recent $R$ positions from score-based selection and selects the $K$ highest-scoring positions from the remaining historical context. The selected historical set is defined as
\begin{equation}
\mathcal{T}_{K}
=
\mathrm{TopK}
\left(
\mathbf{u}_{1:N-R}, K
\right).
\end{equation}

The most recent $R$ positions are always preserved. The final routing set is therefore
\begin{equation}
    \mathcal{G}
    =
    \mathcal{T}_{K}
    \cup
    \{N-R+1,\ldots,N\}.
\end{equation}
When the input length is smaller than $K+R$, all positions are retained.

The same routing mask is applied across Mamba layers while preserving the predefined Global Channels of each layer. Let $\mathcal{C}_{G}$ denote the set of predefined Global Channels and let $\Delta_{t,j}$ denote the original state-update step at context position $t$ and channel $j$. SuPER-Mamba applies the routing mask as

\begin{equation}
\widetilde{\Delta}_{t,j}
=
\begin{cases}
\Delta_{t,j}, & j \notin \mathcal{C}_{G}, \\
\Delta_{t,j}, & j \in \mathcal{C}_{G},\ t \in \mathcal{G}, \\
0, & j \in \mathcal{C}_{G},\ t \notin \mathcal{G}.
\end{cases}
\end{equation}

Thus, for selected context positions, the original update on the Global Channels is retained, whereas unselected context positions do not update these channels. All non-Global Channels remain unchanged. Non-context prompt positions, including the Instruction and Query, bypass Content selection and retain their original updates. SuPER-Mamba therefore modifies long-range state allocation without removing or truncating input tokens, and the complete sequence remains available to the original recurrent computation.

\section{Experimental Setup}
\label{sec:experimental-setup}

\subsection{Models, Baselines, and Datasets}
\label{sec:models-datasets}

\noindent\textbf{Models and Datasets.}
We evaluate SuPER-Mamba on three representative state-space-based models: Mamba1-1.4B~\citep{gu-dao-2023-mamba}, Mamba2-1.3B~\citep{dao-gu-2024-mamba2}, and Zamba2-1.2B~\citep{zamba-2}. All models use their corresponding pretrained parameters without additional training or parameter updates. We evaluate on all 13 tasks in LongBench-E~\cite{bai-etal-2024-longbench}, covering a total of 3,668 test samples.

\noindent\textbf{Baselines.}
We compare SuPER-Mamba with the corresponding base pretrained models and LAMB~\cite{ye-etal-2025-lamb}. For Mamba2, we additionally report LongMamba~\cite{ye-etal-2025-longmamba} to provide a more complete comparison with existing long-context Mamba methods. Since corresponding LongMamba results for Mamba1 and Zamba2 are not included in our evaluation, LongMamba is excluded from the cross-model average.

\noindent\textbf{Evaluation.}
For each LongBench-E task, we compute the official task-specific metric separately over the three context-length ranges of 0--4k, 4--8k, and 8k+. We then take the arithmetic mean of the three range-level scores as the task-level score used in our aggregate results. Overall performance is reported as the arithmetic mean of the 13 task-level scores.

\begin{table}[t]
\centering
\small
\begin{tabular}{lccc}
\hline
\textbf{Category} & \textbf{$K$} & \textbf{Kernel} & \textbf{Pooling} \\
\hline
Single-doc QA     & 1024 & 18 & Max  \\
Multi-doc QA      & 1024 & 10 & Mean \\
Summarization     & 2048 & 18 & Mean \\
Few-shot          & 1024 & 18 & Max  \\
Synthetic         & 256  & 18 & Mean \\
Code              & 1024 & 12 & Mean \\
\hline
\end{tabular}
\caption{Category routing configurations used in the full LongBench-E evaluation. $K$ denotes the number of historical Content positions selected based on the Pooled Surprisal Scores and allowed to update the Global Channels. Kernel denotes the Local Pooling window size $k$, and Pooling denotes Mean or Max aggregation. All task categories additionally retain the most recent $R=32$ Content positions.}
\label{tab:routing-config}
\end{table}

\subsection{Implementation Details}
\label{sec:implementation}

\noindent\textbf{Surprisal Scoring.}
SuPER-Mamba computes Surprisal from the predictive responses of the pretrained model. Context scoring is performed using only the context and does not use the downstream query. The resulting Surprisal scores are fixed before routed inference. No additional selector is trained, and the pretrained model parameters remain unchanged.

\noindent\textbf{Routing Configuration.}
We use predefined Global Channels and apply the selection mask produced by SuPER-Mamba to control Global-Channel updates at different context positions. Token-level Surprisal is first aggregated by local pooling, after which context positions are selected under a fixed routing budget. The most recent 32 context positions are always preserved. We do not apply additional normalization or denoising to the Surprisal scores.

The routing hyperparameters are configured by task category, as summarized in Table~\ref{tab:routing-config}. The dataset-to-category mapping is provided in Appendix~\ref{app:dataset-config}.

\begin{table}[t]
\centering
\small
\begin{tabular}{@{}cccc@{}}
\hline
\textbf{Architecture} & \textbf{Base Model} & \textbf{LAMB} & \textbf{SuPER-Mamba} \\
\hline
Mamba1-1.4B & 8.52 & 12.98 & 17.93 \\
Mamba2-1.3B & 8.09 & 21.30 & 21.56 \\
Zamba2-1.2B & 7.57 & 7.32  & 7.59 \\
\hline
Average & 8.06 & 13.87 & 15.69 \\
\hline
\end{tabular}
\caption{Overall LongBench-E results of SuPER-Mamba on Mamba1-1.4B, Mamba2-1.3B, and Zamba2-1.2B. Each dataset is evaluated using its official LongBench-E task-specific metric, and the results are equally averaged across the three context-length ranges. The Average row reports the arithmetic mean of the overall scores of the three models.}
\label{tab:cross_model_results}
\end{table}

\noindent\textbf{Inference Settings.}
All methods use the same evaluation samples, generation settings, and evaluation protocol. Inference is performed with batch size 1. SuPER-Mamba does not perform input-level context compression. The complete context remains available to the model, while the routing mechanism only controls which context positions update the Global Channels.

\section{Results}

\subsection{Main Results}
\label{sec:main_results}

Table~\ref{tab:cross_model_results} summarizes the overall LongBench-E performance across Mamba1-1.4B, Mamba2-1.3B, and Zamba2-1.2B. Averaged across the three architectures, SuPER-Mamba improves the Base Model average from 8.06 to 15.69, corresponding to a relative improvement of 94.7\%. It also improves over the LAMB average of 13.87 by 13.1\%. The gain is largest on Mamba1-1.4B and Mamba2-1.3B, while the improvement on Zamba2-1.2B is limited. This suggests that predictive feedback can support routing across different Mamba architectures, although its benefit remains dependent on the underlying backbone.
Since LAMB does not provide scripts for Mamba1-1.4B and Zamba2-1.2B, the corresponding results are reproduced based on the method described in the paper.

More fine-grained results on Mamba2-1.3B are reported in Table~\ref{tab:main-results}. SuPER-Mamba outperforms LAMB on 6 of the 13 datasets, with the clearest advantages on Code tasks. LCC improves from 41.65 to 49.03 and RepoBench-P improves from 34.02 to 42.20, corresponding to relative gains of 17.7\% and 24.0\%. Competitive results are also maintained across several remaining datasets, indicating that the overall improvement is not driven by a single task.

\subsection{Ablation Study}
\label{sec:ablation}

Table~\ref{tab:component_ablation} examines the roles of Surprisal and Local Pooling. Raw Surprisal reaches 16.33, improving over the Base Model by 101.9\%, showing that predictive feedback produced by the pretrained model provides an effective signal for long-range position scoring. The Random Control reaches 19.33 under the evaluated setting, indicating that selective allocation of long-range state updates itself can also provide substantial benefits.

Local Pooling further increases the score from 16.33 to 20.85, corresponding to a 27.7\% improvement over Raw Surprisal and a 7.9\% improvement over the Random Control. This result shows that aggregating neighboring predictive responses substantially strengthens Surprisal-based routing. Full SuPER-Mamba further reaches 21.56, demonstrating that predictive feedback, local aggregation, and the complete routing configuration together provide an effective mechanism for long-context state allocation.

\begin{table}[t]
\centering
\small
\begin{tabular}{@{}lc@{}}
\hline
Variant & Average Score \\
\hline
Base Model       & 8.09 \\
Random Control   & 19.33 \\
Raw Surprisal    & 16.33 \\
+ Local Pooling  & 20.85 \\
\hline
Full SuPER-Mamba & 21.56 \\
\hline
\end{tabular}
\caption{SuPER-Mamba component ablation results on Mamba2-1.3B. All results are evaluated on the complete set of 13 LongBench-E datasets using their official task-specific metrics. Each dataset score is obtained by equally averaging the results over the three context-length ranges.}
\label{tab:component_ablation}
\end{table}

\subsection{Hyperparameter Sensitivity}
\label{sec:hyperparameter}

Figure~\ref{fig:hyperparameter_sensitivity} presents the controlled sensitivity analysis of routing budget $K$ and Local Pooling kernel $k$ on a subset of LongBench-E samples. Different task categories exhibit distinct score distributions and best observed settings. Summarization and Synthetic tasks show different responses to the routing budget, while Multi-Document QA and Code show clearer variation across kernel settings. These results indicate that routing budget and local aggregation range affect task categories differently.

Based on these observations, we use category-level configurations rather than a single global setting. For categories requiring finer shared selection, we further perform a restricted joint refinement in which datasets within the same category share one configuration. The selected configurations are then frozen and kept unchanged during the full LongBench-E evaluation. The controlled analysis is therefore used to characterize broad parameter sensitivity, while the subsequent refinement resolves the final shared configuration within a narrower search space. Details of the category-level hyperparameter selection protocol are provided in Appendix~\ref{app:hyperparameter-selection}.

\begin{figure*}[t]
    \centering

    \begin{minipage}[t]{0.485\textwidth}
        \centering
        \includegraphics[width=\linewidth]
        {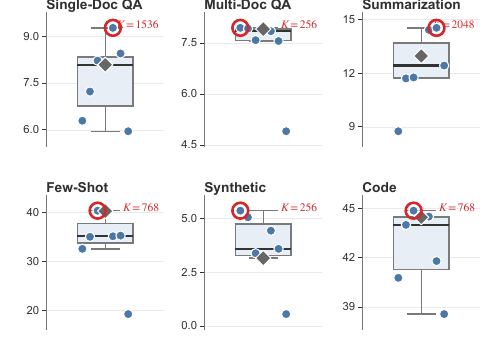}
        \vspace{-1mm}
        \small (a) Routing budget $K$
    \end{minipage}
    \hfill
    \begin{minipage}[t]{0.485\textwidth}
        \centering
        \includegraphics[width=\linewidth]
        {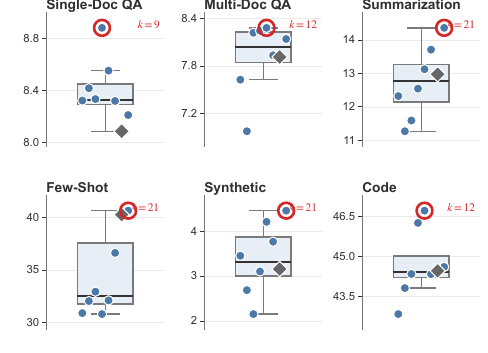}

        \vspace{-1mm}
        \small (b) Local Pooling kernel $k$
    \end{minipage}

    \vspace{-1mm}

    \caption{Hyperparameter sensitivity analysis on a LongBench-E subset using Mamba2-1.3B. The horizontal axis represents the six task categories, and the vertical axis reports the corresponding category-level LongBench-E score obtained by equally averaging the results of the datasets within each category. Each configuration is evaluated on 50 samples from each of the 13 datasets. (a) analyzes the routing budget $K$ with $k=18$ and Mean Pooling fixed, testing $K\in\{256,512,768,1024,1536,2048,3072\}$. (b) analyzes the Local Pooling kernel $k$ with $K=1024$ and Mean Pooling fixed, testing $k\in\{3,5,7,9,12,15,18,21\}$. Boxes summarize the result distributions across all tested settings for each category, gray diamonds indicate the common reference configuration with $K=1024$, $k=18$, and Mean Pooling, and red circles indicate the best observed setting for each category in the corresponding parameter analysis.}
    \label{fig:hyperparameter_sensitivity}
\end{figure*}

\subsection{Length-wise Analysis}
\label{sec:length_analysis}

Table~\ref{tab:lengthwise_results} shows that SuPER-Mamba remains effective as context length increases. The Base Model decreases from 14.98 for inputs from 0 to 4k to 3.85 for inputs above 8k, whereas SuPER-Mamba maintains scores of 23.64 and 19.85 in the same ranges. In the longest range, SuPER-Mamba also exceeds LAMB by approximately 3.7\%. The slower degradation of SuPER-Mamba indicates that selective long range state updates remain effective for longer contexts and help mitigate the loss of earlier information as recurrent processing continues.

\begin{table}[t]
\centering
\small
\begin{tabular}{@{}lccc@{}}
\hline
Method & 0 to 4k & 4 to 8k & 8k+ \\
\hline
Base Model  & 14.98 & 5.45  & 3.85  \\
LAMB        & 22.41 & 22.32 & 19.15 \\
SuPER-Mamba & 23.64 & 21.18 & 19.85 \\
\hline
\end{tabular}
\caption{LongBench-E results of Mamba2-1.3B across different context-length ranges. The 0--4k, 4--8k, and 8k+ columns correspond to the three LongBench-E context-length ranges. Each column uses the official task-specific metric for each dataset and reports the equal average across all 13 datasets.}
\label{tab:lengthwise_results}
\end{table}

\section{Discussion}
\label{sec:discussion}

SuPER-Mamba demonstrates that predictive feedback produced by pretrained Mamba models can serve as an effective control signal for long-context state allocation. By constructing routing scores from Surprisal during context processing, SuPER-Mamba determines long-range state updates without downstream-query-dependent context scoring or additional selector training. The results across Mamba1-1.4B, Mamba2-1.3B, and Zamba2-1.2B suggest that such model-intrinsic predictive signals provide a practical basis for training-free long-context routing across different architectures.


Our experiments show that restricting Global-Channel updates under a fixed budget is beneficial. Performance rises from 8.09 (Base Model, where every position updates the Global Channels; Table~\ref{tab:main-results}) to 19.33 when only a limited number of randomly selected positions are allowed to perform these updates (Appendix Table~\ref{tab:appendix-dataset-ablation}).
This is consistent with the observation that selectively restricting long-range state updates improves performance~\cite{ye-etal-2025-lamb, ye-etal-2025-longmamba}, which raises the question of how much additional value the Surprisal signal provides beyond any reasonable selection policy.

Our results show that Raw Surprisal alone (16.33) underperforms random selection, suggesting that unaggregated predictive responses are noisy. Local Pooling substantially improves the signal (20.85), surpassing the Random Control, and the full configuration reaches 21.56. Thus, while selective allocation accounts for a large fraction of the gains, the combination of Surprisal with local aggregation yields a consistently stronger routing policy than random selection under the same budget. The Surprisal-based scores therefore contribute meaningful structure beyond mere sparsity.

\section{Conclusion}
\label{sec:conclusion}

This paper presents SuPER-Mamba, a Surprisal-guided query-independent routing method for long-context modeling in pretrained Mamba models. SuPER-Mamba constructs positional scores from predictive feedback produced during context processing and applies Local Pooling to aggregate neighboring responses, enabling the allocation of long-range state updates without additional selector training or downstream-query-dependent context scoring. Experiments on LongBench-E show that SuPER-Mamba consistently improves the Base Model and achieves competitive performance across Mamba1-1.4B, Mamba2-1.3B, and Zamba2-1.2B. The ablation study further demonstrates that Surprisal provides an effective predictive signal for long-range routing and that Local Pooling substantially strengthens its effectiveness, while the length-wise analysis shows that the method remains effective under longer context conditions. Overall, SuPER-Mamba demonstrates that predictive feedback from pretrained models can serve as an effective signal for long-range information routing, providing a training-free approach to long-context modeling in Mamba.

\section{Limitations}
\label{sec:limitations}

SuPER-Mamba currently uses fixed task-category-specific routing configurations, which cannot adapt to differences across individual inputs. In addition, the offline design requires an extra context-scoring pass before routed inference, introducing additional computation and latency. Future work could explore input-adaptive routing and more efficient score computation while preserving the query-independent and training-free properties of the framework.

\bibliography{custom}
\clearpage

\appendix

\section{Dataset Categories and Routing Configurations}
\label{app:dataset-config}

We group the 13 LongBench-E datasets into six task categories. Datasets within the same category share the selection budget $K$, Local Pooling kernel size $k$, and pooling operator. All category-level configurations are determined before the full LongBench-E evaluation and remain fixed throughout the main experiments. Table~\ref{tab:appendix-dataset-config} provides the complete dataset-to-category mapping and the corresponding routing configurations. All configurations use a fixed recent-position window of $R=32$.

\begin{table*}[t]
\centering
\small
\setlength{\tabcolsep}{7pt}
\renewcommand{\arraystretch}{1.15}

\begin{tabular}{llccc}
\hline
\textbf{Dataset}
& \textbf{Category}
& \textbf{$K$}
& \textbf{Kernel}
& \textbf{Pooling} \\
\hline
Qasper & Single-doc QA & 1024 & 18 & Max \\
MultiFieldQA-en & Single-doc QA & 1024 & 18 & Max \\
HotpotQA & Multi-doc QA & 1024 & 10 & Mean \\
2WikiMQA & Multi-doc QA & 1024 & 10 & Mean \\
GovReport & Summarization & 2048 & 18 & Mean \\
MultiNews & Summarization & 2048 & 18 & Mean \\
TREC & Few-shot & 1024 & 18 & Max \\
TriviaQA & Few-shot & 1024 & 18 & Max \\
SAMSum & Few-shot & 1024 & 18 & Max \\
Passage Count & Synthetic & 256 & 18 & Mean \\
Passage Retrieval-en & Synthetic & 256 & 18 & Mean \\
LCC & Code & 1024 & 12 & Mean \\
RepoBench-P & Code & 1024 & 12 & Mean \\
\hline
\end{tabular}
\caption{Complete mapping from LongBench-E datasets to task categories and their corresponding category routing configurations. Different categories use different routing parameter configurations while following the same routing mechanism during inference.}
\label{tab:appendix-dataset-config}
\end{table*}

\begin{table}[t]
\centering
\small
\setlength{\tabcolsep}{4.5pt}
\renewcommand{\arraystretch}{1.12}

\begin{tabular}{lccc}
\hline
\textbf{Category}
& \textbf{$K$}
& \textbf{Kernel $k$}
& \textbf{Pooling} \\
\hline
Single-doc QA & 1024 & 18 & Max \\
Multi-doc QA & 1024 & 10 & Mean \\
Summarization & 2048 & 18 & Mean \\
Few-shot & 1024 & 18 & Max \\
Synthetic & 256 & 18 & Mean \\
Code & 1024 & 12 & Mean \\
\hline
\end{tabular}
\caption{Final category routing configurations used for the full LongBench-E evaluation. $K$ denotes the selection budget for historical Content positions, Kernel $k$ denotes the Local Pooling window size, and Pooling denotes Mean or Max aggregation. These configurations are determined from the parameter analysis on a LongBench-E subset and remain fixed throughout the subsequent full evaluation.}
\label{tab:appendix-hyperparameter-config}
\end{table}

\section{Detailed Routing Implementation}
\label{app:routing-implementation}

SuPER-Mamba follows an offline two-pass inference procedure. The first pass constructs the routing scores and the corresponding routing mask, while the second pass performs routed inference using the fixed mask. The complete prompt is preserved throughout both passes. SuPER-Mamba does not remove or truncate Content positions and instead controls which positions are allowed to update the predefined Global Channels.

During the first pass, Surprisal scoring is performed only on the Content and does not use the downstream Query. Position-wise predictive responses are computed for the Content and aggregated by Local Pooling to obtain the routing scores. These scores are then associated with their corresponding Content positions in the complete prompt. Instruction, Query, and other non-Content prompt positions do not participate in the Content Top-$K$ selection and therefore do not compete for the selection budget.

For Content routing, the most recent $R=32$ positions are excluded from score-based historical selection and are always retained. Among the remaining historical Content positions, SuPER-Mamba selects the $K$ positions with the highest pooled scores. The final Content routing set therefore consists of the historical Top-$K$ positions together with the Recent-$R$ positions. When the Content length does not exceed $K+R$, all Content positions are retained.

The resulting Content mask is then projected to the complete prompt. All non-Content positions retain their original updates. Selected Content positions retain their original updates on the Global Channels, whereas unselected Content positions are prevented from updating only these channels. Their remaining recurrent computation is unchanged. The same position-level routing mask is applied to the predefined Global Channels across Mamba layers. During the second pass, the model receives the complete Instruction, Content, and Query, while the routing mask obtained from the first pass remains fixed. This implementation keeps context scoring independent of the downstream Query while preserving the full input sequence during routed inference.

\section{Surprisal and Local Pooling Implementation Details}
\label{app:surprisal-pooling}

\noindent\textbf{Surprisal Alignment.}
During the first pass, SuPER-Mamba tokenizes the Content independently and performs a forward pass using the pretrained Mamba without routing modification. For each position $t>1$, the logits produced at position $t-1$ are compared with the ground-truth token at position $t$ using cross-entropy, so the resulting Surprisal score is aligned with the predicted token position rather than the position producing the logits. The first Content token has no preceding prediction within the independently processed Content sequence and is therefore assigned a Surprisal score of zero. Logits may be computed in chunks to reduce memory usage without changing the resulting position-wise Surprisal scores.

\noindent\textbf{Projection to the Complete Prompt.}
Because independently tokenizing the Content does not necessarily produce token boundaries identical to those of the same Content span inside the complete prompt, SuPER-Mamba does not assume a direct one-to-one correspondence between the two tokenizations. The implementation first identifies the exact character span of the Content in the rendered prompt and obtains tokenizer offset mappings for both the Content-only sequence and the complete prompt. Prompt tokens overlapping the Content span are then identified. For each such prompt token, Surprisal scores from overlapping Content-only tokens are combined using a character-overlap-weighted average. If no direct character overlap is available, the score of the nearest Content token by character-center distance is used. This projection produces one routing score for each Content position in the complete prompt.

\noindent\textbf{Local Pooling.}
For a given kernel size $k$, SuPER-Mamba applies stride-1 Mean Pooling or Max Pooling to aggregate Surprisal responses over neighboring positions, using padding of $k/2$ according to the pooling implementation. For even kernel sizes such as $k=18$, the underlying pooling operation produces one additional output position. The final extra position is therefore removed so that the pooled score sequence has the same length as the original Content position sequence. The resulting Pooled Surprisal Scores are subsequently used for Content routing.

\begin{table*}[t]
\centering
\small
\setlength{\tabcolsep}{5.5pt}
\renewcommand{\arraystretch}{1.12}

\begin{tabular}{lcccc}
\hline
\textbf{Dataset}
& \textbf{Random Control}
& \textbf{Raw Surprisal}
& \textbf{+ Local Pooling}
& \textbf{Full SuPER-Mamba} \\
\hline
Qasper                 & 4.12  & 3.36  & 4.85  & \textbf{5.00} \\
MultiFieldQA-en        & 13.00 & 11.70 & \textbf{14.40} & 13.81 \\
HotpotQA               & 6.54  & 5.71  & \textbf{6.82}  & 6.43 \\
2WikiMQA               & 8.83  & 8.39  & 10.00 & \textbf{10.09} \\
GovReport              & 12.58 & 10.65 & \textbf{13.80} & 13.70 \\
MultiNews              & 12.89 & 11.85 & 14.31 & \textbf{14.45} \\
TREC                    & 29.33 & 5.33  & 39.67 & \textbf{40.33} \\
TriviaQA                & 42.14 & 39.89 & 44.82 & \textbf{47.69} \\
SAMSum                  & 25.11 & 26.67 & 27.84 & \textbf{28.30} \\
Passage Count           & 3.90  & 4.22  & \textbf{4.29}  & 4.00 \\
Passage Retrieval-en    & 4.63  & 4.73  & 4.25  & \textbf{5.19} \\
LCC                     & 44.80 & 40.32 & 43.94 & \textbf{49.03} \\
RepoBench-P             & \textbf{43.38} & 39.44 & 42.07 & 42.20 \\
\hline
\textbf{Average}        & 19.33 & 16.33 & 20.85 & \textbf{21.56} \\
\hline
\end{tabular}

\caption{Dataset-level component ablation results on LongBench-E using Mamba2-1.3B. Each dataset is evaluated using its official LongBench-E task-specific metric. Samples are divided into the 0--4k, 4--8k, and 8k+ context-length ranges, and the three range-level scores are averaged arithmetically to obtain the dataset-level score. The Average row reports the arithmetic mean over all 13 dataset-level scores. Bold values indicate the best result among the four component settings.}
\label{tab:appendix-dataset-ablation}
\end{table*}

\section{Hyperparameter Selection Protocol}
\label{app:hyperparameter-selection}

SuPER-Mamba adopts shared routing configurations at the task-category level. The 13 LongBench-E datasets are first grouped into six categories according to their task type and input-context structure: Single-doc QA, Multi-doc QA, Summarization, Few-shot, Synthetic, and Code. This categorization is defined independently of the subsequent hyperparameter experiments. Datasets within the same category share the same routing budget $K$, Local Pooling kernel size $k$, and pooling operator, avoiding dataset-specific hyperparameter optimization.

After defining the task categories, we analyze the effects of the routing budget $K$, Local Pooling kernel size $k$, and pooling operator using Mamba2 on a subset of LongBench-E. We adopt a one-variable-at-a-time parameter sweep with $K=1024$, $k=18$, and Mean Pooling as the common reference configuration. For the routing budget, we fix $k=18$ and Mean Pooling and evaluate

\begin{equation}
K \in \{256,512,768,1024,1536,2048,3072\}.
\end{equation}

For the Local Pooling kernel, we fix $K=1024$ and Mean Pooling and evaluate

\begin{equation}
k \in \{3,5,7,9,12,15,18,21\}.
\end{equation}

The pooling operator is evaluated by comparing Mean and Max Pooling under $K=1024$ and $k=18$. All 13 LongBench-E datasets are included in the parameter analysis, with 50 samples evaluated for each dataset under each configuration. The shared reference configuration is reused across the individual sweeps, resulting in 15 distinct configurations in total.

Hyperparameter selection is performed within the predefined task categories. For each category, we jointly consider the performance of its constituent datasets under different configurations and select a shared setting that maintains consistently strong performance across the category. As reported in Section~\ref{sec:hyperparameter}, the sensitivity analysis aggregates the dataset-level scores within each category using an equal average to characterize the effect of individual hyperparameters. The final configurations are determined from the experimental results at the category level rather than by applying a single automatic optimization rule.

The resulting category-level routing configurations are summarized in Table~\ref{tab:appendix-hyperparameter-config}. These configurations are fixed before the full LongBench-E evaluation and remain unchanged throughout the subsequent experiments.

\section{Additional Experimental Results}
\label{app:additional-results}

This section provides additional dataset-level and length-wise results for the component ablation study reported in Section~\ref{sec:ablation}. All results are obtained on Mamba2 using the same LongBench-E evaluation protocol and task-category-specific routing budgets as in the main experiments.

\subsection{Dataset-level Component Ablation}
\label{app:dataset-ablation}

Table~\ref{tab:appendix-dataset-ablation} reports the dataset-level results for Random Control, Raw Surprisal, Surprisal with Local Pooling, and Full SuPER-Mamba across all 13 LongBench-E datasets. The four variants are evaluated under the same task-category-specific routing budgets, allowing the effects of different routing signals and aggregation components to be compared at the dataset level.

The dataset-level results show that the effects of individual routing components vary across tasks. Raw Surprisal substantially improves overall performance over the Base Model reported in the main text, but its effectiveness is not uniform across datasets. Introducing Local Pooling increases the average score from 16.33 to 20.85 and improves over Raw Surprisal on 12 of the 13 datasets. This result provides further evidence that aggregating neighboring predictive responses substantially strengthens Surprisal-based routing.

Full SuPER-Mamba further increases the average score to 21.56 and outperforms the Local Pooling variant on 9 of the 13 datasets. The strongest variant is not identical across every individual task, indicating that the contribution of the complete routing configuration can vary with task characteristics. Nevertheless, the aggregate improvement is distributed across multiple task categories rather than being attributable to a single dataset.

\subsection{Length-wise Component Results}
\label{app:lengthwise-ablation}

Table~\ref{tab:appendix-length-ablation} further reports the component ablation results across the three LongBench-E context-length ranges. For each range, the reported score is the arithmetic mean over the same 13 datasets.

\begin{table}[t]
\centering
\small
\setlength{\tabcolsep}{5pt}
\renewcommand{\arraystretch}{1.12}

\begin{tabular}{lcccc}
\hline
\textbf{Variant}
& \textbf{0--4k}
& \textbf{4--8k}
& \textbf{8k+}
& \textbf{Average} \\
\hline
Random Control      & 21.45 & 18.51 & 18.03 & 19.33 \\
Raw Surprisal       & 19.17 & 15.87 & 13.94 & 16.33 \\
+ Local Pooling     & 23.05 & 20.45 & 19.06 & 20.85 \\
Full SuPER-Mamba    & \textbf{23.64} & \textbf{21.18} & \textbf{19.85} & \textbf{21.56} \\
\hline
\end{tabular}

\caption{Length-wise component ablation results on LongBench-E using Mamba2-1.3B. The 0--4k, 4--8k, and 8k+ columns correspond to the three context-length ranges. Each range uses the official LongBench-E task-specific metric for each dataset and reports the equal average across all 13 datasets. Bold values indicate the best result in each column.}
\label{tab:appendix-length-ablation}
\end{table}

The length-wise results are consistent with the overall ablation findings. Raw Surprisal performs below the Local Pooling variant across all three context-length ranges. Local Pooling increases the average scores to 23.05, 20.45, and 19.06 for the 0--4k, 4--8k, and 8k+ ranges, respectively, indicating that the benefit of local aggregation is not restricted to a particular input-length regime.

Full SuPER-Mamba achieves the highest average score in all three ranges, reaching 23.64, 21.18, and 19.85, respectively. In the 8k+ range, performance increases from 13.94 with Raw Surprisal to 19.85 with the complete method. These results further indicate that combining predictive feedback with Local Pooling and the complete routing configuration provides more robust long-context performance than position-level Surprisal alone.

\end{document}